# Localized Partial Evaluation of Belief Networks

Denise L. Draper  &  Steve Hanks
Department of Computer Science and Engineering
University of Washington, Seattle, WA 98105
{ddraper, hanks}@cs.washington.edu

## Abstract

Most algorithms for propagating evidence through belief networks have been *exact* and *exhaustive*: they produce an exact (point-valued) marginal probability for every node in the network. Often, however, an application will not need information about every node in the network nor will it need exact probabilities. We present the *localized partial evaluation* (LPE) propagation algorithm, which computes *interval* bounds on the marginal probability of a specified query node by examining a subset of the nodes in the entire network. Conceptually, LPE ignores parts of the network that are "too far away" from the queried node to have much impact on its value. LPE has the "anytime" property of being able to produce better solutions (tighter intervals) given more time to consider more of the network.

## 1 Introduction

Belief networks provide a way of encoding knowledge about the probabilistic dependencies and independencies of a set of variables in some domain. Variables are encoded as nodes in the network, and relationships between variables as arcs between nodes. Algorithms such as Pearl's polytree propagation algorithm [Pearl, 1988] and Lauritzen and Spiegelhalter's clustering algorithm [Lauritzen and Spiegelhalter, 1988] can solve the network, where "solving the network" means computing the exact marginal probability of each of the nodes in the network, possibly conditioned on evidence (observations of the values of some of the variables). In general, this problem becomes computationally intractable as the network size increases [Cooper, 1990]. While in practice existing algorithms often suffice for small or carefully crafted networks, we are interested in the prospect of using large, automatically-generated networks to encode such things as an entire outcome space for a planner or a projector, e.g. [Kushmerick et al., 1994], [Dean and Kanazawa, 1989]. Furthermore, it is often the case that we are really only interested in the marginal probability of a small number of variables, and perhaps wish to know only bounds on the probabilities (*e.g.* to within a certain tolerance, or relative to a threshold)—for example, we might decide to execute a plan if it achieves a certain goal with probability greater than 85%. Solving an entire network may be a very inefficient way to obtain bounds on the probabilities of a few variables.

*Localized partial evaluation* (LPE) is a new network evaluation algorithm which incrementally refines interval bounds on the marginal probability of a given query node. Other incremental bounding algorithms ([Horvitz et al., 1989], [D'Ambrosio, 1993],[Poole, 1993]) produce bounds by considering only some cases (possible node instantiations) over the entire network. In contrast, LPE produces bounds by considering only a subset of the nodes in the network.

LPE is based on standard message-propagation techniques, but uses interval-valued messages instead of point-valued messages, and performs its calculations over only a subset of the nodes in the network. Briefly, when given a query for a particular node, some *active subset* of the nodes in the network is identified. Messages which would, in a standard algorithm, have been sent from nodes outside the active subset are replaced with *vacuous messages*, composed of [0, 1] intervals. The interval value of a vacuous message captures the range that message could have taken on had it actually been computed (*i.e.* had the message's source been in the active subset). The nodes inside the active subset compute messages in the normal fashion, but they are now interval-valued messages, due to the influence of the vacuous messages received from outside the active subset. Finally, interval bounds on the marginal probability of the query node are computed from the interval-valued messages.

By iterating over successively larger active subsets, LPE can produce successively narrower interval bounds. In the limit, when all variables relevant to the query node (*i.e.* not d-separated from the query node) have been included in the active set, LPE generates the true point-valued marginal probability. The rate of convergence is not guaranteed; in the worst case, many iterations may be needed to produce sufficiently narrow intervals. Since iterating over successive active subsets may require recomputing messages multiple times, LPE can be slower than simply solving the entire network. On the other hand, the expected size of the active subset needed to answer a query (and thus the time required to answer it) need not in principle grow *at all* with the size of the network.

The outline of the rest of this paper is as follows: Section 2 defines the localized partial evaluation algorithm for singly-connected networks. Section 3 extends the algorithm to multiply-connected networks. Section 4 presents preliminary empirical results, Section 5 discusses related work, and Section 6 concludes.

## 2 Localized Partial Evaluation of Polytrees

LPE operates by selecting a subset of the nodes and arcs in the network, known as the *active set*, and running a message-propagation algorithm—such as Pearl's polytree propagation algorithm—over this set. Messages which would have been received over arcs not in the active set are replaced by *vacuous messages*: vectors of [0, 1] intervals. Owing to the influence of these vacuous messages, messages computed within the active set may also be (and generally are) interval-valued. The three components of this algorithm—computation of the interval-valued messages, choice of an active set, and propagation of messages over this set—are largely independent, and we will discuss each in turn.

### 2.1 Computing Interval-Valued Messages

We use intervals to denote bounds on the possible value of some quantity. Thus, for example, a vector of [0, 1] intervals indicates that the true value of each position in the vector lies somewhere between 0 and 1, but we don't know where (because we haven't computed it yet). In order to compute $\lambda$ or $\pi$ messages with interval-valued inputs, we must develop *interval extensions* of the $\lambda$ and $\pi$ functions.

There are two criteria for an interval extension of a function: (1) the interval (or interval vector) must be correct (contain the true value), and (2) the interval should be of minimum width. The simple arithmetic functions have well-known interval extensions, e.g. $[a, b] + [c, d] = [a + c, b + d]$. Any function constructed out of these arithmetic extensions is guaranteed to be correct, but it is *not* guaranteed to be minimum width. Consider, for example, the equation $m + n = [\underline{m} + \underline{n}, \overline{m} + \overline{n}]$ ($\underline{m}$ denotes the lower bound of $m$ and $\overline{m}$ the upper). What if $m$ and $n$ are known to vary inversely? That is, in those situations in which $m$ takes on its minimum value, $n$ must take on its maximum, and vice versa? In that case, neither the lower bound of $\underline{m} + \underline{n}$ nor the upper bound of $\overline{m} + \overline{n}$ can ever occur. As an extreme example, the interval equation $m - m$ does not produce $[0, 0]$, as we would like, but rather $[\underline{m} - \overline{m}, \overline{m} - \underline{m}]$. This problem occurs whenever inputs to a computation are dependent. To reiterate, in interval computations, independence is a *conservative* assumption, producing answers that are guaranteed to be correct, but liable to be wider than necessary (unless the inputs truly are independent).

The message computations used in polytree computation consist primarily of additions and multiplications, which can easily be extended to their interval equivalents. However since each $\pi$ or $\lambda$ message is a vector of dependent intervals, such a direct extension typically produces unacceptably wide results. The *Annihilation/Reinforcement algorithm* for computing $\pi$ and $\lambda$ messages [Tessem, 1992, Tessem, 1989] can produce narrower results. The crux of the A/R algorithm is a technique for obtaining narrower bounds on any computation of the form $X = \sum_i a_i b_i$, where $a$ and $b$ are interval-valued vectors, and $b$ is known to represent a normalized quantity (that is, the 'true' values $b'_i$ obey the constraint $\sum_i b'_i = 1$). Typically, $\sum_i \underline{b_i} < 1$ and $\sum_i \overline{b_i} > 1$, so evaluating $X$ at these extrema produces wider bounds than necessary. Instead, we want to pick $b^*_i \in b_i$ such that $\sum_i b^*_i = 1$ and $X$ is minimized or maximized. $\underline{X}$ is computed by first initializing $b^*_i = \underline{b_i}$, then considering the $\underline{a_i}$ in sorted ascending order, and successively incrementing the corresponding $b^*_i$ to $\overline{b_i}$ until the limit $\sum_i b^*_i = 1$ is reached. Finally, $\underline{X} = \sum_i \underline{a_i} b^*_i$. $\overline{X}$ is found similarly, but sorting $\overline{a_i}$ in descending order.

The A/R algorithm is quite efficient when the length of the vectors is small. When $i$ is large, the cost of sorting $a$ becomes expensive. We have implemented the A/R algorithm with an incremental QuickSort which reduces this cost by sorting only as much of $a$ as needed.

We will use $A/R_i(a_i, b_i)$ to indicate the A/R computation of $\sum_i a_i b_i$, and $\hat{BEL}$, $\hat{\pi}$ and $\hat{\lambda}$ to indicate the interval valued extensions of BEL, $\pi$ and $\lambda$ messages. Then the equations for the (unnormalized) $\hat{BEL}$, $\hat{\pi}$ and $\hat{\lambda}$ using the A/R algorithm are:

$$\hat{\pi}(x) = \sum_u \left( P(x|u) \prod_i \hat{\pi}_X(u_i) \right)$$

$$= A/R_u \left( P(x|u), \prod_i \hat{\pi}_X(u_i) \right)$$

$$\hat{\lambda}(x) = \prod_i \hat{\lambda}_Y(x)$$

$$\hat{BEL}(x) = \hat{\lambda}(x)\hat{\pi}(x)$$

$$\hat{\pi}_{Y_j}(x) = \hat{\pi}(x) \prod_{k \neq j} \hat{\lambda}_{Y_k}(x)$$

$$\hat{\lambda}_X(y) = \sum_k \left( \hat{\lambda}(x_k) \sum_{u \setminus y} \left( P(x_k|u, y) \prod_i \hat{\pi}_X(u_i) \right) \right)$$

$$\begin{aligned}
&= \sum_{u \backslash y} \left( \sum_k \left( \hat{\lambda}(x_k) P(x_k | u, y) \right) \prod_i \hat{\pi}_X(u_i) \right) \\
&= \underset{u \backslash y}{A/R} \left( \underset{k}{A/R} \left( P(x_k | u, y), \hat{\lambda}(x_k) \right), \prod_i \hat{\pi}_X(u_i) \right)
\end{aligned}$$

The second argument to the A/R algorithm must be normalized, thus both $\pi$ and $\lambda$ messages must be normalized in the above equations.[1]

These equations do not produce the theoretically narrowest possible bounds: the normalization operation the equations use does not produce minimum width results when the elements to be normalized are dependent (which they are in this case). [Tessem, 1992] discusses the difficulties of producing true minimum-width results. Our experience shows that in practice these equations produce sufficiently narrow results (and in particular, they converge rapidly as the input messages converge).

### 2.2 Propagating Messages

We use a standard polytree propagation algorithm, as defined in [Pearl, 1988], with a a few minor modifications for efficiency. Since we are only computing the belief value for a single node, messages only need to be computed in a single direction, thus our propagation makes only one pass over the network instead of two. We cache computed messages. If the algorithm is iterated, and a message computation produces the same value as the cached value, it does not need to be propagated a second time. Iteration in polytrees is thus relatively efficient.

### 2.3 Choosing the Active Set

An active set consists of some connected subset of the nodes and arcs in a belief network. (It is sometimes more convenient to think of the active set as consisting only of nodes; in polytrees the two representations are interchangeable, but when we consider multiply-connected networks, which arcs are included in the active set will matter.) When a query is initially directed at a node, an active set is constructed containing only the query node. Thereafter, a loop ensues: computation is done over the current active subset; if the result is not satisfactory (according to some user-supplied criteria, *e.g.* interval-width or threshold), a larger active set is computed and the process repeats. Thus the problem we are faced with is how best to extend an active set given that its current boundaries are inadequate.

Our research in this area is still preliminary. On our randomly generated polytrees, a simple breadth-first strategy has worked better than any other strategy we have tried. The breadth-first strategy extends the active set at each iteration by including its directly neighboring nodes, except that nodes known to be d-separated from the query node are not added. In general, two factors affect whether a node should be added to the active set: the impact that information from this node will have on the query answer (its relevance), and the cost of computing that information. In principle, we could use decision-theoretic techniques to (greedily) add nodes in the order of highest expected gain per additional unit of computing time. In these terms, the breadth-first strategy uses distance from the query node as a measure of relevance, and ignores cost entirely. Future work will explore more sophisticated measures of relevance based on the nodes' conditional probabilities and the current message values, as well as considering computational cost.

## 3 Multiply-Connected Networks

There are at least three ways in which the LPE algorithm defined in the last section might be extended to multiply-connected networks:

1. Use a clustering method which creates a tree structure, such as [Lauritzen and Spiegelhalter, 1988], and apply LPE to the resulting tree.

2. For each multiply-connected portion of the network (*knots*, in the terminology of [Peot and Schachter, 1991]), insist that either all or none of the nodes and arcs composing the knot be in the active set. Use an interval-valued extension to an algorithm for multiply-connected networks, such as conditioning or clustering, on the knots in active set. The cost associated with evaluating any knot outside the active set is eliminated.

3. Place no *a priori* limitations on the contents of the active set. Rather, given an active set, determine which knots are *wholly contained* in the active set, and use an interval-valued extension to an algorithm for multiply-connected networks on just those knots. Portions of knots partially contained in the active set are treated as singly-connected portions of the network, with vacuous messages sent over all arcs not in the active set (see Figure 1).

The first two of these possibilities are relatively straightforward to implement. Presumably, most techniques for evaluating knots could be extended to handle interval-valued messages using the A/R algorithm, as long as the normalization constraints of the A/R algorithm can be met. We have extended the technique of *cutset conditioning*, [Pearl, 1988, Suermondt and Cooper, 1991], modifying the algorithm in two ways: In order to condition over individual knots instead of over the entire network [Peot and Schachter, 1991], $\lambda$ messages originating from outside the knot must be treated as indirect evidence. Secondly, the probability of each cutset instance $C_i$ is interval-valued. By definition, $\sum_i P(C_i) = 1$, therefore we use the A/R algorithm to compute the final probability $\sum_i P(X|C_i)P(C_i)$. Unfortunately, doing so requires storing the entire set of $P(X|C_i)$ and $P(C_i)$ values, which means that our implementation requires space exponential in the size of the cutset, negating one of the main advantages of the conditioning algorithm. Further work is required to explore the benefits and difficulties of interval extensions of other evaluation algorithms.

---

[1] Normalization is another operation where the obvious interval extension would produce unnecessarily wide results. Instead, an interval vector $b$ is normalized for each position $i$ in $b$ by setting $\underline{b_i} := \underline{b_i}/(\underline{b_i} + \sum_{j \neq i} \overline{b_j})$, and $\overline{b_i} := \overline{b_i}/(\overline{b_i} + \sum_{j \neq i} \underline{b_j})$.

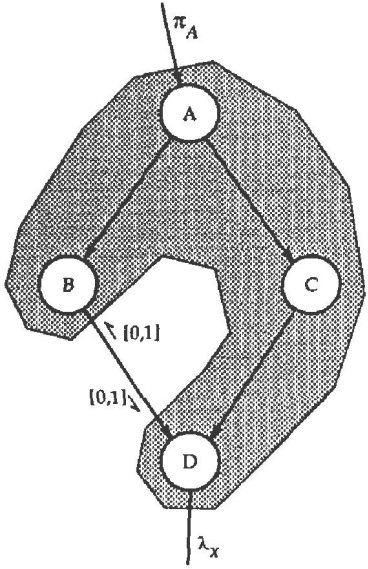

Figure 1: The active set does not include the whole loop.

The last strategy for extending LPE to multiply-connected networks—using polytree messages over singly-connected parts of knots—is not obviously correct. Let us call an arc that is excluded from the active set, but which connects two nodes that are in the active set, a *missing arc* (*e.g.* the arc from B to D in Figure 1). Then the argument that this strategy is correct consists of these two claims: (1) in the equation for the marginal (or conditional) probability of any node in the active set, the presence of a missing arc has the effect of replacing the summation over the states of the source of the arc (*e.g.* B) with an interval taking the minimum and maximum over the states of the source of the arc,[2] and (2) when normalized, this interval expression contains the true value.

Proving the first claim requires an inductive proof which we will not give here; instead we will demonstrate how the summation is replaced with the interval in the derivation of the expression for $\hat{BEL}_C(c)$ in Figure 1: The true expression we must correctly bound is

$$BEL_C(c) = \sum_{a,b,d,y} P(a|y)P(b|a)P(c|a)$$
$$P(d|b,c)\pi_A(y)\lambda_X(d) \quad (1)$$

---
[2]Expressions which include, but do not sum over, states of B (*e.g.*, $BEL_B$ and $\pi$ messages sent from B to its children), are simply vacuous, and thus trivially contain the true result.

The principal 'trick' we use is that $A/R_i(a,[0,1]) = [\min_i a_i, \max_i a_i]$; for compactness, we will denote this interval by $mm_i\, a$. Then, computing the messages propagated toward the node C:

$$\hat{\lambda}_B(a) = \sum_b P(b|a)\hat{\lambda}_D(b)$$
$$= A/R_b(P(b|a), [0,1])$$
$$= mm_b P(b|a)$$

$$\hat{\pi}_C(a) = \hat{\pi}(a)\hat{\lambda}_B(a)$$
$$= \sum_y (P(c|y)\hat{\pi}_C(y))\, mm_b P(b|a)$$

$$\hat{\lambda}_D(c) = \sum_b \left(\sum_d (P(d|b,c)\hat{\lambda}_X(d))\, \hat{\pi}_D(b)\right)$$
$$= A/R_b \left(\sum_d (P(d|b,c)\hat{\lambda}_X(d)), [0,1]\right)$$
$$= mm_b \sum_d P(d|b,c)\hat{\lambda}_X(d)$$

$$\hat{BEL}_C(c) = \hat{\lambda}_D(c)\sum_a P(c|a)\hat{\pi}_C(a)$$
$$= mm_b \sum_d P(d|b,c)\hat{\lambda}_X(d)$$
$$* \sum_a \left(P(c|a)\right.$$
$$\left.\sum_y P(c|y)\hat{\pi}_C(y)\, mm_b P(b|a)\right)$$
$$= mm_b \sum_{a,d,y} P(a|y)P(b|a)P(c|a)$$
$$P(d|b,c)\hat{\pi}_A(y)\hat{\lambda}_X(d) \quad (2)$$

Thus we have replaced a summation over $b$ in Equation (1) by an interval containing the min and max over $b$ in Equation (2).[3]

The second claim is simpler to demonstrate: what we wish to show is that when we replace $\sum_b$ with $mm_b$

---
[3]We made several simplifications in this derivation for clarity; namely we left out the normalization of the messages, and we aggregated two occurences of $mm_b$ into one in the last step. The effect of both of these simplifications is to narrow the result, *i.e.* the interval given by Equation (2) is narrower than the interval actually computed by LPE. Since we will show that Equation (2) contains the true belief $BEL_C$, it follows that the wider interval computed by LPE must also contain $BEL_C$.

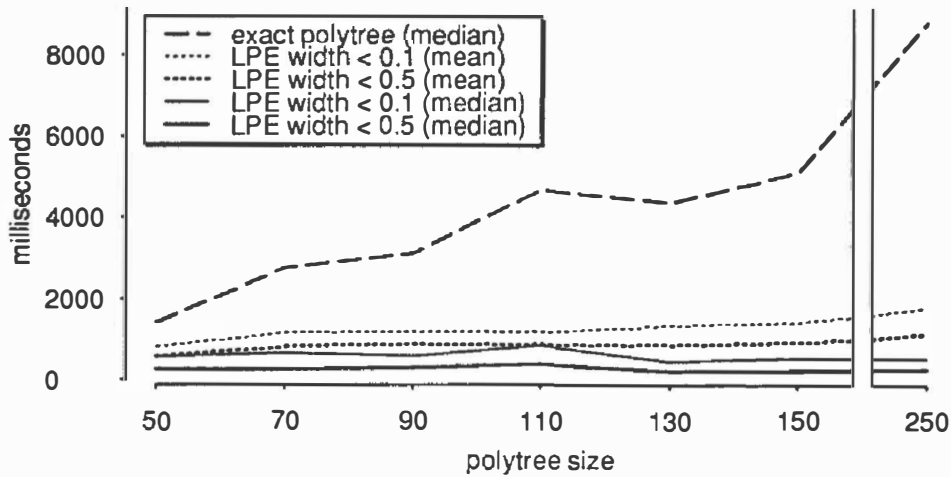

Figure 2: Performance on polytrees with increasing network size.

and normalize the results, the interval expression contains the former expression. Assuming that the variable that we are normalizing over is C, and that the number of states of C is $n$, we can write the normalized summation expression as $\sum_b f(b,c)/\sum_{b,c} f(b,c)$ and the lower normalized bound as:

$$\frac{\min_b f(b,c)}{\min_b f(b,c) + \sum_{c' \neq c} \max_b f(b,c')}$$

$$= \frac{n \min_b f(b,c)}{n \min_b f(b,c) + \sum_{c' \neq c} n \max_b f(b,c')}$$

$$= \frac{\sum_b f(b,c) - \delta}{(\sum_b f(b,c) - \delta) + (\sum_{b,c' \neq c} f(b,c') + \delta')}$$

$$\leq \frac{\sum_b f(b,c)}{\sum_{b,c} f(b,c)},$$

since $(x-\delta)/(x-\delta+y) \leq x/(x+y)$ for positive $x, y, \delta$. The upper bound is shown similarly.

Using LPE on incomplete knots has the potential to be quite powerful. If the active set is chosen in such a way as to avoid including entire knots, then we can use polynomial algorithms on sections of the network where we would otherwise be forced to use exponential algorithms. In the extreme, we might adopt a "no-loops" strategy, requiring that the active set *never* add arcs which would complete loops, thus guaranteeing that the active set is always a polytree. Of course, LPE might not be able to refine the query interval sufficiently without taking into account the missing arcs. Still, even if it were relatively unlikely to succeed, the computational advantage of the polytree algorithm might make it worthwhile to pursue a no-loops strategy as an initial attempt to answer a query.

## 4  Empirical Results

We have implemented localized partial evaluation on top of the IDEAL belief net package [Srinivas and Breese, 1990], written in Allegro Common Lisp, running on a Sun Workstation. In this section we will present some performance results using LPE to evaluate queries on randomly generated networks. In these networks, nodes could have between two and four states, and the number of parents of each node was effectively limited by limiting the size of the conditional probability table to not more than 1000 values. Node probabilities were randomly set according to a skewed distribution[4] which is intended to more closely represent the sort of distributions one would expect to see in real networks. (Extreme conditional probabilities also present more of a challenge for LPE, since they increase the potential dependence between nodes.) Evidence was generated for a random number of nodes (maximum 1/4 of the network). All times in this section measured are in milliseconds, using the AllegroCL "time" function, and taking the non-gc user time.

### 4.1  Polytrees

We generated polytrees with sizes ranging from 50 to 250 nodes. For each non-evidence node in each network, we iterated LPE until the width of the belief interval was no greater than a target width (0.5 or 0.1). For comparison, we also ran the polytree algorithm on each tree. In total, 85 trees were generated, and tests performed on over 10000 nodes. We experimented with different techniques for expanding the active set between iterations, but simple breadth-first expansion out-performed our other tests, so that is the only technique presented here.

---

[4] For each probability value an integer between 1 and 10 and an exponent between 1 and 5 are selected independently, and appropriate sets of these values are normalized together. The result is a distribution which tends to be skewed towards very small and large values; the "expected skew" is around two orders of magnitude.

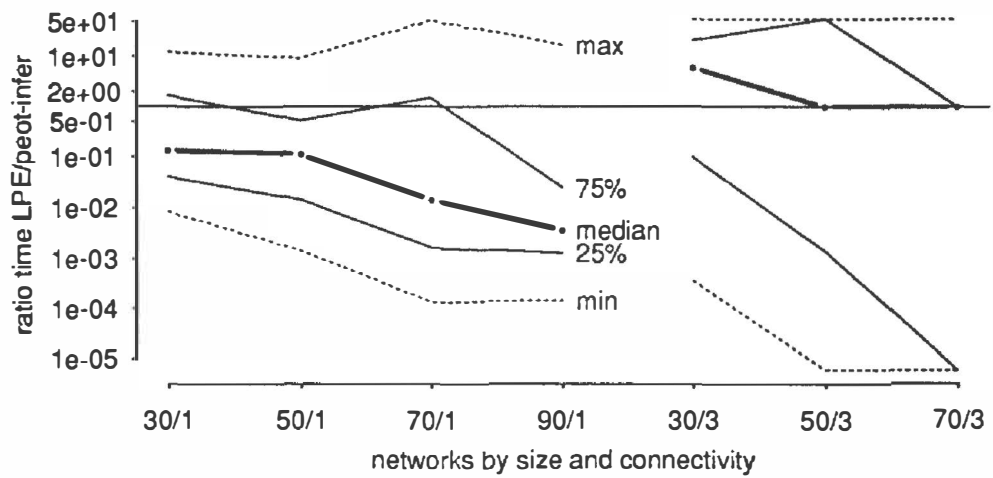

Figure 3: Performance of LPE on multiply-connected networks with a query width of 0.1. The $x$-axis shows individual networks categorized by type, e.g. the category "50/3" contains networks with 50 nodes and 1.3 arcs/node.

Figure 2 shows that LPE performed extremely well on these tests. The median time to respond to a query remains nearly flat with increasing network size, and the mean grows only slightly. Analysis of the data shows further that while the number of nodes in the largest active set necessary to answer a query (i.e. the active set of the final iteration) grows somewhat with increasing network size, the number of iterations required to answer the query remains constant. Thus the rise in the mean with larger network size is most likely an artifact of the increased connectivity of the larger networks—that is, the time to answer a query appears to be independent of the *diameter* of the network.

LPE performs satisfactorily in the worst case as well. LPE never required more than 1.6 times as much time to answer a query than the polytree algorithm. Finally, LPE outperformed the polytree algorithm 97.5% of the time (with the ratio improving as the tree size increased).

### 4.2 Multiply-Connected Networks

In our first attempts to test multiply-connected networks, we used the random network generator provided in IDEAL. The resulting networks were highly connected (an average of twice as many arcs as nodes), and nearly impossible to evaluate by any method. So we turned to sparser networks. We generated a series of polytrees with between 30 and 90 nodes, and then randomly added arcs until we had achieved a particular ratio of arcs to nodes (1.1 and 1.3). The networks with a 1.3 ratio were constructed by adding arcs to the networks with a 1.1 ratio, thus the 1.3 networks are strictly "harder" than the 1.1 networks. As in the polytree experiments reported above, nodes could have between two and four states, distributions were set randomly according to a skewed distribution, and evidence was generated for up to one fourth of the nodes in the network. For each network, fifteen randomly-selected nodes were tested.

The first experiments concern the no-loops active set extension strategy discussed at the end of Section 3. We implemented the no-loops strategy as a variant of the breadth-first strategy (so the arcs excluded from the active set are the last arcs of each loop to be encountered in the breadth-first expansion); future work includes investigating whether there are heuristics to determine which arcs should be excluded. But even this simple strategy performed reasonably well. We queried each of the test nodes, requesting an interval of zero width, thus forcing LPE to produce the narrowest interval it could with this strategy. We found no correlation between interval width and network size, but Figure 4 shows that there is correlation between interval width and the connectedness ratio of the network (1.1 and 1.3): the sparser networks converge better. For the sparser 1.1 networks, LPE was able to obtain intervals of width less than 1/2 for 76% of the nodes; for the 1.3 networks, 40% of the nodes.

To test the behavior of LPE's extension of loop conditioning, we implemented another variant on the breadth-first strategy: expansion is breadth-first, except that arcs which would connect loops are delayed for several (five) rounds. This is an arbitrary strategy, and we have not yet experimented with any others, so this data should be regarded as extremely preliminary. We evaluated the test nodes requesting interval widths of 0.1. For comparison, we evaluated the networks with peot-infer, an efficient version of cutset conditioning supplied with IDEAL.

The times for both peot-infer and LPE vary by sev-eral orders of magnitude, even for networks of the same size, so we show instead the ratio of their performance: for each query node, the time for LPE to answer the query is compared to the time for peot-infer to solve that network. Figure 3 shows that in sparser (1.1) networks, LPE performs progressively better relative to peot-infer as the network size increases. For the 1.3 networks, the data is inconclusive. LPE performance is slower than peot-infer for the 3❶ node networks, about even for the 50 node networks, and for the 70 node networks most of the tests of both LPE and peot-infer timed out, making comparison meaningless.[5]

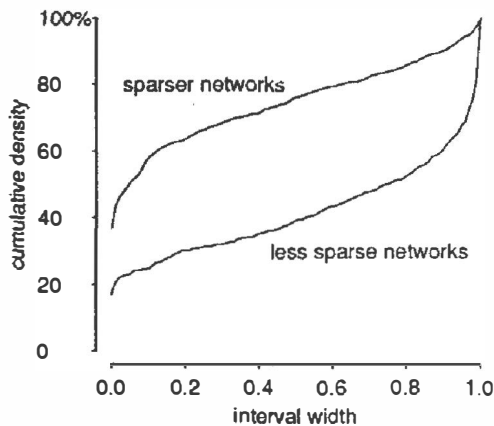

Figure 4: Width of intervals using polytree algorithm on multiply-connected networks.

Our results show that LPE performs very well in singly-connected networks, and can perform well on sparse multiply-connected networks. Our work is still preliminary however, particularly with respect to strategies for choosing the active set. Also, randomly-generated networks may not reflect the structure of real networks, particularly for multiply-connected networks, so we hope to extend this work by applying LPE to real networks.

## 5 Related Work

There is a long history of computation with interval-valued probabilities or degrees of belief (including, *e.g.* [Fertig and Breese, 1989], [Kyburg, 1987], [Shafer, 1976], [Dubois and Prade, 1990]). The intent of these works is to capture the notion of "upper and lower" degrees of belief—that is, degrees of belief are taken to be intrinsically interval-valued. This is intended by some to be fundamental stand on how degrees of belief should be represented, and by others as a practical technique for eliciting information from experts. In either case, our work differs in that the source of inter-vals under partial evaluation is purely a *computational* artifact. One important ramification of this difference is the definition of a correct answer: if intervals are taken to be the fundamental representation of degree of belief, then there is a single correct, interval-valued, answer to any query. Under localized partial evaluation any interval that contains the true point-valued probability is a correct answer.[6]

Others, *e.g.* [Quinlan, 1983], [Thöne *et al.*, 1992], have pursued intervals as bounding approximations, but usually in the context of rule bases rather than networks. [Hanks and McDermott, 1994] bounds queries with respect to a threshold in dynamically-constructed networks of a highly restricted form, also relying on the assumption that a single evidence source will be sufficient to answer the query.

Three works similar in spirit to LPE are bounded conditioning [Horvitz *et al.*, 1989], (incremental) SPI [D'Ambrosio, 1993, Li and D'Ambrosio, 1992], and the search-based algorithm of [Poole, 1993]. All three are anytime algorithms which incrementally refine bounds on a solution by taking into account progressively more information, and all three attempt to hasten convergence by processing information in order of greatest impact on the solution. Bounded conditioning is a version of cutset conditioning which incrementally processes cutset instances, producing interval results by bounding the impact of all as-yet-uncomputed cutset instances. SPI is an inference algorithm which constructs a factored evaluation tree to efficiently compute probabilities expressed in the form of Equation (1). SPI is made incremental by computing larger terms of each factor first, and constructing an error bound on the possible remaining probability mass. Poole's search algorithm operates similarly, except that instead of factoring the expression, it generates the most likely complete instances (assignments to all nodes) using a top-down search. Thus all three of these algorithms acquire partial information by considering cases incrementally, and exploit the skewness of the joint probability distribution for convergence.

LPE, in contrast, acquires partial information by considering parts of the network incrementally. These two sources of partial information are complementary; LPE will in fact perform better the *less* skewed the conditional distributions are. This suggests that it may be possible to combine LPE with one of the other algorithms. Using an interval-valued extension of SPI to evaluate knots in LPE is one promising possibility.

## 6 Conclusion

Localized partial evaluation is a new algorithm for belief network propagation which incrementally refines interval bounds on the marginal probabilities of indi-vidual nodes. LPE generates bounds by considering only a subset of the nodes in a network, unlike previous algorithms which have produced bounds by considering a subset of node instantiations (cases). LPE can be used on both singly- and multiply-connected networks. A novel feature of LPE is its ability to bound probabilities by using a polytree propagation algorithm over subsets of multiply-connected components of a network.

---

[5]In Figure 3, timed-out data points are placed at the maximum or minimum of the graph according to whether LPE or peot-infer timed out, unless both peot-infer and LPE timed out, in which case they are placed at 1.0.

[6]LPE could easily be extended to provide bounds on interval-valued probabilities; indeed the A/R algorithm was intended for that purpose.

## Acknowledgements

We wish to thank the reviewers for very helpful comments on this work. This work is supported by NASA GSRP Fellowship NGT-50822 and by NSF grant IRI-9206733.